\documentclass[
twocolumn,
]{ceurart}

\begin{document}

%%
%% Rights management information.
%% CC-BY is default license.
\copyrightyear{2021}
\copyrightclause{Copyright for this paper by its authors.
  Use permitted under Creative Commons License Attribution 4.0
  International (CC BY 4.0).}

%%
%% This command is for the conference information
\conference{CDCEO 2022: 2nd Workshop on Complex Data Challenges in Earth
Observation, July 25, 2022, Vienna, Austria}

%%
%% The "title" commandhttps://www.overleaf.com/project/627e0e0357b9770a3a47db01
\title{Multimodal Crop Type Classification Fusing Multi-Spectral Satellite Time Series with Farmers Crop Rotations and Local Crop Distribution}

%%
%% The "author" command and its associated commands are used to define
%% the authors and their affiliations.
\author[1]{Valentin Barriere}[%
orcid=0000-0002-0877-7063,
email=valentin.barriere@ec.europa.eu,
%url=https://yamadharma.github.io/,
]
\address[1]{European Commission's Joint Research Center, Via Fermi, 2749, 21027 Ispra VA, Italy}

\author[1]{Martin Claverie}[%
orcid=0000-0002-0877-7063,
email=martin.claverie@ec.europa.eu,
%url=https://yamadharma.github.io/,
]
% \author[1]{Anonymous}[%
% orcid=0000-0002-0877-7063,
% email=Anonymous@Anonymous,
% %url=https://yamadharma.github.io/,
% ]
% \address[1]{Anonymous}

% \author[1]{Anonymous}[%
% orcid=0000-0002-0877-7063,
% email=Anonymous@Anonymous,
% %url=https://yamadharma.github.io/,
% ]

%%
%% The abstract is a short summary of the work to be presented in the
%% article.
\begin{abstract} 
Accurate, detailed, and timely crop type mapping is a very valuable information for the institutions in order to create more accurate policies according to the needs of the citizens. In the last decade, the amount of available data dramatically increased, whether it can come from Remote Sensing (using Copernicus Sentinel-2 data) or directly from the farmers (providing in-situ crop information throughout the years and information on crop rotation).
Nevertheless, the majority of the studies are restricted to the use of one modality (Remote Sensing data or crop rotation) and never fuse the Earth Observation data with domain knowledge like crop rotations. Moreover, when they use Earth Observation data they are mainly restrained to one year of data, not taking into account the past years. 
In this context, we propose to tackle a land use and crop type classification task using three data types, by using a Hierarchical Deep Learning algorithm modeling the crop rotations like a language model, the satellite signals like a speech signal and using the crop distribution as additional context vector. We obtained very promising results compared to classical approaches with significant performances, increasing the Accuracy by 5.1 points in a 28-class setting (.948), and the micro-F1 by 9.6 points in a 10-class setting (.887) using only a set of crop of interests selected by an expert. We finally %conducted extensive analysis of our results in order to understand and propose possible applications of our model. 
proposed a data-augmentation technique to allow the model to classify the crop before the end of the season, which works surprisingly well in a multimodal setting. 
\end{abstract}
%\vspace*{-2cm}
%%
%% Keywords. The author(s) should pick words that accurately describe
%% the work being presented. Separate the keywords with commas.

%%
%% This command processes the author and affiliation and title
%% information and builds the first part of the formatted document.
\maketitle

\vspace*{-1.2cm}
%%%%%%%%% BODY TEXT
\section{Introduction} \vspace*{-.2cm}

Timely and accurate crop type mapping provides valuable information for crop monitoring and productions forecast \cite{becker2010monitoring}. In-season crop type mapping can serve not only to better estimate the crop areas, but also to improve the yield forecasting by using crop-type specific models. Crop type mapping is thus a major information of the crop monitoring systems focusing to in-season forecast of the crop production.

The high-spatial resolution time series enables to determine crop type at a sub-parcel level in most agricultural areas. Most of the remote sensing classification systems relies on supervised techniques, requiring in-situ crop identification survey. If the survey data are provided within the season, some systems \cite{defourny2019near} are designed to predict crop type along the season with a given uncertainty, even if the crop cycle is on-going; %it is named in-season crop type mapping (or early crop mapping). However, 
such surveys data are expensive because of the need of labels from the current year to train a model, difficult to achieve at large scale and in most cases delivered after the cropping season. 
There is a high demand for %in-season 
crop type mapping that does not rely on survey data from the on-going season. Such approaches, as the one proposed in this study, are based on model trained with past seasons and applied on the current one, plus we proposed a data-augmentation method to obtain satisfying results earlier in the season.

\paragraph*{Earth Observation-based crop type mapping} \vspace*{-.3cm}
Machine learning classification methods have been widely tested to derive crop type map from remote sensing data. Among the various methods, Random Forest algorithm has proved its capacity to accurately identify crop type, accounting for large and non parametric data set \cite{hansen1996classification}. Since 2015 and the launch of the first satellite of the Copernicus Sentinel-2 (S2) constellation, the perspective for crop type mapping at large scale  has changed. The high spatial and temporal resolution of S2 offers indeed an appropriate data set to distinguish crop type, based on the spectral and temporal signals, at parcel or sub-parcel level in most agricultural region. Taking benefit of this capacity, some operational systems have been expended  \cite{inglada2015assessment,defourny2019near,johnson20102009}, combining Earth Observation (EO) data, in situ observations and classifier algorithm to deliver crop type maps at regional, country scale or continental scale \cite{d2021parcel}.

% https://www.sciencedirect.com/science/article/pii/S0034425717304686
\paragraph*{Crop type mapping using Deep-Learning method} \vspace*{-.3cm}
The recent progresses in deep-learning benefit the crop type mapping applications. In \cite{Russwurm2019}, the authors are classifying crop types at the parcel-level, using the data from the French Brittany during the season 2017. The authors have compared a Transformer-Encoder \cite{Vaswani2017} and a Recurrent Neural Network of type Long-Short-Term-Memory (LSTM) \cite{Hochreiter1997}. They obtain comparable results between the Transformers and the LSTM, %the former obtaining the best accuracy (0.69) and the latter the best macro-F1 (0.59). 
obtaining best accuracy (0.69) for the former and macro-F1 (0.59) for the latter. 

In \cite{Russwurm2020}, the authors have designed a crop classifier at the parcel-level using S2 and compared several approaches to model the signal, comprising a Transformer and a LSTM. They obtain respective overall accuracies between 0.85 and 0.92 using the LSTM depending on the number of classes considered. 
A similar approach has been run by \cite{Russwurm2019b} 
%In \cite{Russwurm2019b}, the authors have run similar approach 
on 40k Central Europa parcels using S2. They proposed a new early classification mechanism in order to enhance a classical model with an additional stopping probability based on the previously seen information. 
%By adding an earliness reward loss to the main loss function coupled with a special weighting of the global loss which is based on the output stopping probability, they build a model that predicts better and faster the 7 crop types. 

Finally, \cite{Russwurm2018} are using the same technique developed in \cite{Rubwurm2017}, where they tackle the task of crop classification at the pixel level, i.e. accounting for the spatial variation to detect parcels boundaries. They are using a CNN-LSTM network on S2 images to classify 17 types of crops. 

%To put in contrast \cite{Russwurm2019,Russwurm2020,Russwurm2019b} with our work, 
%the data they use to test their models is from the same year than the one they are using for training, which means that they need to have to wait to get labels from the test year to train their model. They also use more spectral bands. Our work finally differs from them by the fact we are processing multimodal information, by taking into account the crop rotations over the years, and the local crop %distribution of the parcel neighborhood. 

\vspace*{-.2cm}
\paragraph*{Modeling the crop rotation sequences}
Crop rotation is a widely-used agronomic technique for sustainable farming, preserving the long term soil quality. Good understanding and design of crop rotation are essentials for sustainability and to mitigate the variability of agricultural productivity induced by climate change.  %\cite{BOHAN2021169}
The crop rotation depends on the farmer management decision, but some good practices are shared, enabling to model the crop rotation patterns \cite{dogliotti2003rotat}. They remains nonetheless complex and non stable in time; changes may be related to, e.g. economic consideration (commodities price) or administrative regulation (e.g. subsidies changes). Expert knowledge based models are thus very limited and rarely accurate over large areas and long periods. Alternatively, estimation of the crop sequence probabilities without a priori using survey data and hidden Markov models has been demonstrated in France \cite{xiao2014modeling}. However, survey data are not always available. Relying on machine learning techniques, \cite{Osman2015} use a Markov Logic model in order to predict the following year's crop in France, with an accuracy of 60\%. In \cite{Yaramasu2020}, the authors focused on deep deep neural networks to reach a maximum accuracy of 88\% on a 6-class portion of the US Cropland Data Layer (CDL) dataset over 12 years \cite{Boryan2011}.

%\paragraph*{Crop Varieties (copy paste)}
%Knowledge of crop variety can also help to reduce negative impacts associated with disease spread, providing important information for certification and control. For example, the identification of varieties using data from an orbital-borne sensor could help institutions that breed varieties for royalty charges for the propagation of their genetic material, by reducing evaluation time and field-checking efforts (cf. the Brazilian Cultivars Protection Law [176]). Finally, varieties can be essential to the definition of a geographical indication, like in vineyards where the inventories maintained by government authorities from information passed on by producers could, to some degree, be certified using remote sensing data [177].

\vspace*{-.2cm}
\paragraph*{Motivation}

A lot of works are focusing on the use of remote sensing to predict the crop type at pixel or parcel level using only the EO and in-situ observations of the current year. Nevertheless, they consider the signal as independents from a year to another. 
Other works are using the crop rotations of the parcels in order to tackle a pre-season prediction of the crop type, focusing on a few classes problem. In this case, it is obvious there is too much information missing to reach high performances.
%Finally, the use of local and stable information like the crop distribution of the region is crucial to give strong context during the prediction of the crop.  
As of 2022, we identified a single study combining the use of crop rotations and satellite time-series data over several years: \cite{johnson2021pre}. They present a methodology to derive near real time Cropland Data Layer over major US agricultural states. The methodology is nonetheless restricted to a limited number of crop types and the use of Random Forest classifier, while the recent progress in deep learning shows tremendous improvements in such data mining problem. 
%We added the crop distributions related to the field over that?  

\vspace*{-.2cm}
\paragraph*{Contributions}

We propose to model both the crop rotations and the S2 time series signal in a multimodal way using a hierarchical Long-short-term-memory (LSTM). The contribution is unique in term of conception as no work has been proposed fusing the large amount of temporally fine-grained EO data with crop rotation analysis in an advanced deep learning method. The crop rotations and the S2 time series were enhanced by the use of the crop distributions of the neighborhood fields picked from previous year. 
%We compared to another state-of-the-art approach and obtain significantly better results.
The crop rotations are modeled over the year as words would be in a language model \cite{mikolov2010recurrent}, helped by the S2 time-series data that are modeled as if it was the prosody of the speaker. Finally, the high-level features we add on the last layer of the network could be seen as the distribution of the words used by our speaker. 
Finally, we also propose a data-augmentation technique for the in-season classification, by randomly cropping the end of the RS time-series data. It allows to learn a model able to classify the type of crop without the whole time-series, hence before the end of the season. 

\begin{figure*}[!t]
    \centering 
    %\hspace*{-1cm}
    \includegraphics[width=1.\textwidth]{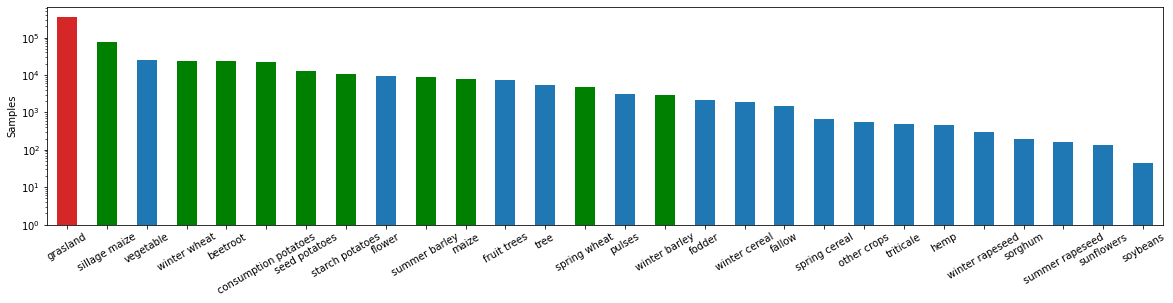} \vspace*{-.5cm}
    \caption{Distributions of the crop types in the dataset. %Blue crops are aggregated into other crops for the 12-class setting
    Green crops are the remaining ones for the 10-class evaluation.}
    \label{fig:distribcrop} \vspace*{-.5cm}
\end{figure*} 

%% FIG TO IMPROVE
% \begin{figure*}[!t]
%     \centering 
%     %\hspace*{-1cm}
%     \includegraphics[width=1.\textwidth]{img/SchemaDraft.png}
%     \caption{General overview of the study combining satellite earth observation, crop rotation and spatial crop distribution as inputs for the LSTM model. The training data is selected from 2016 to 2019 for EO and from  2009 to 2019 for crop rotation. The trained model is then validated over 2020.}
%     \label{fig:Generalscheme} \vspace*{-.5cm}
% \end{figure*} 

\vspace*{-.2cm}
\section{Methodology} \vspace*{-.2cm}

% \subsection{Learning Model}

\subsection{Dataset} \label{subsec:data}

The study is focused on data acquired over The Netherlands, covers the period 2009-2020 for the crop type labeling and the parcel identification, and the period 2016-2020 for the S2 data.

\vspace*{-.2cm}
\subsubsection*{Crop Type data}

The crop type data were obtained from the Dutch Land Parcel Identification System and GeoSpatial Aid Application, named \textit{Basis registratie Percelen} (BRP). % managed by the Netherlands Enterprise Agency (\textit{Rijksdienst voor Ondernemend Nederland}). 
Dutch  farmers  must  annually  record  their  field  parcel  boundaries  and  associated cultivated  crops.%  The   annual   datasets   can   be   accessed   online.
\footnote{https://data.overheid.nl/data/dataset/basisregistratie-gewaspercelen-brp}
The 12 yearly BRP (2009-2020) were merged through geographical polygon intersections. The output polygons correspond to the 12-year intersected areas and there are associated with 386 crop codes. The polygons which areas are lower than half an hectare were discarded. The output product %(named hereafter the Multiyear BRP Stack, MBS) 
contains 974,000 polygons covering a total of 1,600 Mha.%\footnote{as a comparison, the 2019 BRP covers 1,870 Mha with 772,500 parcels}

%The data set includes 386 distinct cultivated code items. Knowing, the focus of this study on crop types, a set of land use and land cover categories was aggregated by an expert from the domain.
% a set of land use and land cover categories was aggregated and selected with respect to the aims and objectives of the min- istry.
%Hence, we propose several aggregation levels yielding to 28 and 12 classes as shown in Figure \ref{fig:distribcrop}.

For the evaluation, we propose 3 granularities of labels, using several aggregations lead by an expert from the domain and yielding to 386, 28 and 12 crop classes.

\vspace*{-.2cm}
\subsubsection*{Sentinel-2 Data}

\paragraph*{Data}

The study relies on the analysis of the optical Copernicus Sentinel-2 (S2) data. S2 constellation provides observations with a minimum revisit of five days over ten land spectral bands of the optical domain (460-2280 nm), with a spatial resolution of 10-20 meters depending on the bands. The data are processed up to surface reflectance (SR) Level 2A accounting for atmospheric corrections and cloud/cloud-shadow screening using sen2cor algorithm \cite{louis2016sentinel}. The data are available though the JEODPP platform \cite{soille2017jrc}. Cloud free SR data were processed to 20-m Leaf Area Index (LAI) and Fraction of Absorbed Photosynthetically Active Radiation (FAPAR) using BV-NET \cite{baret2007lai} and calibration settings of \cite{claverie2013validation}. For each polygon, B4 (red band) SR, B8A (near infrared band) SR, LAI and FAPAR were averaged at polygon level using pixels  in a 20-m inner buffer in order to remove parcel edge effects.

\vspace*{-.2cm}
\paragraph*{Time series Smoothing}

Despite the cloud and cloud-shadow screening of L2A S2 products, noise  remains in the resulting time series \cite{claverie2018harmonized}. We applied a time series outliers detection based on B4 (for omitted cloud) and B8A (for omitted cloud-shadow) and using the Hampel filer \cite{pearson2002outliers}. Filtered data were removed for the four variables.
The filtered time series of the four variables were smoothed using the Whittaker algorithm \cite{eilers2017automatic} implemented by the World Food Program.\footnote{https://github.com/WFP-VAM/vam.whittaker} %Whittaker smoothing algorithm is simple to implement massively and is well adopted by the remote sensing community. 
Time series were first resampled and interpolated to a 2-day time step and then the Whittaker algorithm using the V-curve optimization of the smoothing parameter is applied. It yielded to 2-day smoothed time series of each of the four variables, from October N-1 to October N for cropping season of year N.
%The time series were finally resampled to 15-day time step using a 30-day moving window. We considered data from October-1rst of the previous year to October-31st of the current year. Therefore the time series were composed by a total of 25 data points. 

\vspace*{-.2cm}
\subsection{Feature Extractions}

\subsubsection*{Crop Types}

The crop types labels %from the BRP (see Subsection \ref{subsec:data})  
contains 386 different types of crops over the 12 years of study. We model the crop by a one-hot vector of size $V=386$ and used it as an input to an embedding layer. 

The crop label categories for 2020, the year  used as test set, correspond to a long-tailed class distributions, as shown for the 28-class aggregation in Figure \ref{fig:distribcrop}.

\vspace*{-.2cm}
\subsubsection*{EO-based Features}

We integrate the EO time series spatially by averaging at the parcel-level, then temporally using a sliding window of size 30 days and a step size of 15 days. For each parcel, this yields to 25 windows for the whole year for each of the Remote Sensing (RS) signal, that we integrated temporally using 7 statistical functionals: mean, standard deviation, 1st quartile, median, 3rd quartile, minimum and maximum. In total we obtain 7*4=28 features per window, leading to 700 features per year.

With this configuration we have overlap between the windows and avoiding to loose information by breaking the signal dynamics, at the price of a bit of redundancy in the features. On each window, we integrated each signal using statistical functionals like it would be done for speech data \cite{Schuller2016}.

\vspace*{-.2cm}
\subsubsection*{Spatial Crop Distribution}
The spatial crop distribution was derived for the year 2019 (year N-1 as compared to the 2020 validation test set). For each polygon, we compute the sum of the surface for each crops of the data base included in a 10-km circle and turned it to percentage. This a-priori distribution of crops is proven to be relatively stable in time with minor change from year to year \cite{merlos2020scale}. We round the probability at $10^-4$, leading to some values being 0 when not null. 

\vspace*{-.2cm}
\subsection{Learning Model} \vspace*{-.1cm}

This section describes the learning model and the  the features' integration as observations.

\vspace*{-.2cm}
\subsubsection*{Unimodal RNN-LSTM Crop Rotations model}

% \paragraph*{Unimodal Crop Rotations model}

% The most simple layer of modelization in our network is the crop rotation. 
We are modeling the crop rotation at the level of a year by using a LSTM that is trained like a language model. Indeed, it is possible to see each crop like a token in a sentence and train a recurrent neural network that will learn to predict the next word regarding the preceding words. 

We firstly add an embedding layer to transform the crop type $c_t$ at time $t$ into a vector (see Equation \ref{eq:emb}). 
\begin{equation}
    \textbf{emb}_t = f_e(c_t)
    \label{eq:emb}
\end{equation}

Then we feed this vector into the RNN to produce a hidden state $h_t$ at time $t$ (see Equation \ref{eq:rnn}), which will be used to predict the next crop $c_{t+1}$ (see Equation \ref{eq:cropmodel}).

\begin{equation}
    \textbf{h}_t = LSTM_y(\textbf{emb}_t|\textbf{h}_{t-1})
    \label{eq:rnn}
\end{equation}

%\begin{equation}
%    h_t = LSTM(emb(c_t), emb(c_{t-1}), ..., emb(c_{1}))
%    \label{eq:rnn}
%\end{equation}

\begin{equation}
    P(c_{t+1}|c_{t}, ..., c_{1}) = f_c(\textbf{h}_t)
    \label{eq:cropmodel}
\end{equation}

% \paragraph*{Multimodal model with RS}

\vspace*{-.2cm}
\subsubsection*{Features from RS signal}

Using only the past rotations to predict the following year's crop is very difficult, hence we chose to add available information from satellite data in order to make the model more robust.  

Firstly, we enhance the unimodal LSTM crop model by adding information from RS and aligned it at the year-level before concatenating the unimodal RS vector with the crop embedding. %In order to align the RS at the year-level w
Secondly, we chose to process the RS signal beforehand using another RNN and concatenated this unimodal RS vector obtained with the crop embedding, in a Hierarchical way. Those networks are denoted with a Hier- in their name. 

\paragraph*{Multimodal model with RS}

For the first model, we integrated the RS features at the year-level before the LSTM modeling the crop types. We feed the 700 features $RS_t$ into a neural network layer $f_{rs}$ to reduce their size and then concatenate them with the crop embeddings before the LSTM (see Equation \ref{eq:MMRNN}), using $emb_{MM_t}$ instead of $emb_t$ in Equation \ref{eq:rnn}. This model denoted as $\text{LSTM}_{MM}$

\begin{equation}
    \textbf{emb}_{MM_t} = [\textbf{emb}_t, f_{rs}(\textbf{RS}_t)]
    \label{eq:MMRNN}
\end{equation}

\vspace*{-.2cm}
\paragraph*{Bidirectional RNN-LSTM with attention to model the RS time-series}

The first model presented above does not take into account the sequentiality of the RS signal. We decided to correct this aspect by processing the RS features at the year level with a first RNN before adding their yearly representation into the second neural network modeling the crop types, leading to a hierarchical network \cite{Serban2015}. This will give 28 features per window $\textbf{RS}_{t_w}$, for a sequence length of 25 per year. 

We chose to enhance a simple LSTM with a bidirectional LSTM (biLSTM) with a self-attention mechanism \cite{Bahdanau2016} following the assumption that some parts of the year are more important than others to discriminate the crop type. This model denoted as $\text{HierbiLSTM}_{MM}$

The biLSTM is composed of 2 LSTM, one of each read the sequence forward and the other reads it backward. The final hidden states are a concatenation of the forward and backward hidden states. For a sequence of inputs $[\textbf{RS}_{t_1},...,\textbf{RS}_{t_w}]$ it outputs $w$ hidden states $[\textbf{h}_{RS_{t_1}},..., \textbf{h}_{RS_{t_w}}]$. The attention layer will compute the scalar weights $u_{t_w}$ for each of the $\textbf{h}_{RS_{t_w}}$ (see Equation \ref{eq:att}) in order to aggregate them to obtain the final state $\textbf{h}_{RS_t}$ 
(see Equation \ref{eq:blstm}).

\begin{equation} \vspace*{-.2cm}
    u_{t_w} = att(\textbf{h}_{RS_{t_w}})
    \label{eq:att}
\end{equation}

\begin{equation} \vspace*{-.2cm}
    \textbf{h}_{RS_t} = \sum\limits_{w} u_{t_w} \textbf{h}_{RS_{t_w}}
    \label{eq:blstm}
\end{equation}

% \subsubsection*{Bidirectionnal RNN-LSTM with Attention to model the radio time-series}

\vspace*{-.2cm}
\subsubsection*{Locally aggregated crop distributions}

When classifying at the scale of a whole country, the agricultural practices like the type of crops that are used can change. Typically the distribution of the crop types in a region is a stable value over the years and represent the kind of crops supposed to be found in this part of the world. We integrated this local information by adding a vector representing the distributions over the crop types in an area corresponding to a circle of 10-km centered around the studied parcel. 

We chose to add the distribution vector before the last layer because it is a high-level feature regarding the task we are tackling and the deeper you go into the layers the higher-level the representations are w.r.t. the task \cite{Sanh2017}. We concatenated the hidden state $\textbf{h}_t$ of the LSTM with the crop distribution vector $\textbf{d}$ and mixed them using two fully connected layers $f_{fc1}$ and $f_{fc2}$ (see Equation \ref{eq:distrib}). Hence, we obtain $\textbf{h}_{d_t}$ instead of $\textbf{h}_{t}$ before the final fully connected layer $f_{fc}$ from Equation \ref{eq:cropmodel}. This final model is denoted as Final. 

\begin{equation} \vspace*{-.2cm}
    \textbf{h}_{d_t} = f_{fc2}(f_{fc1}([\textbf{h}_t, \textbf{d}]))
    \label{eq:distrib}
\end{equation}

\vspace*{-.2cm}
\section{Experiments and Results}

In this section we will describe the different experiments and results we ran with all the different models. Because of the nature of our predictions, it can be useful to get them before the end of the farming season. In this context, we ran experiments using different setups when predicting, we used an end-of-season configuration and an early-classification configuration. For the end-of-season configuration we feed the neural network with all the RS data of the year while in the early-classification configuration we stop to different date of the year. We compared using LSTM processing the RS data and tagging at the year-level, seeing all the year in an independent way. This year-independent model obtained state-of-the-art results according to \cite{Russwurm2019} and is denoted as $\text{LSTM}_{YI}$.

%We present results for 3 granularities of labels, by using several aggregations lead by an expert from the domain and yielding to 386, 28 and 12 (10 crops only) classes.

% We compared 

\begin{table*}[!t]
\centering
\resizebox{\textwidth}{!}{
\begin{tabular}{l|c|llll|llll|llll|llll}
 \textbf{Labels} & \multirow{2}{*}{\textbf{\#Modal.}}  &  \multicolumn{4}{c}{\textbf{386-class}} & \multicolumn{4}{c}{\textbf{28-class}} & \multicolumn{4}{c}{\textbf{12-class}} & \multicolumn{4}{c}{\textbf{10-class}} \\ %\hline
  \textbf{Model} & & P & R & F1 & Acc &  P & R & F1 & Acc &  P & R & F1 & Acc &  P & R & F1 & m-F1 \\  \hline \hline 
$\text{LSTM}_{Crop}$    &1   (C)     & 28.1  & 22.4  &  23.1  &  73.3   & 45.3 & 33.4 & 34.2  & 76.4 & 53.1 & 44.7 & 43.8 & 77.2 & 46.7& 39.2 & 37.1 & 52.1 \\\hline
$\text{LSTM}_{YI}$ \cite{Russwurm2019}   & 1 (RS)        & 14.3  & 9.8  &  9.9 &  72.5   & 53.1 & 45.1 & 45.9  & 88.5 & 75.0 & 66.3 & 67.7 & 90.4 & 72.1 & 62.2 & 63.7 & 80.3 \\
$\text{LSTM}_{RS}$    &1     (RS)      &  13.7 & 11.8  & 10.8  &   72.5  & 49.7  & 47.0 & 44.1 & 87.4 & 70.6 & 69.4 & 65.3 & 89.0 & 67.3 & 65.3 & 60.7 & 76.0 \\ 
$\text{HierbiLSTM}_{RS}$   &1    (RS)        &  10.7 & 10.0  &    9.0 &  78.7   &  48.0 & 48.1  & 44.5 & 88.7 & 72.7 & 71.0 & 67.7 & 90.7 & 69.3 & 66.6 & 63.0 & 79.1 \\ \hline
$\text{LSTM}_{MM}$   &2  (RS+C)  & 32.5  &26.1   & 26.2   &  86.8  & 63.3 & 57.4 & 56.7 & 91.8 & 79.9 & 78.5 & 78.2 & 93.2 & 77.9&75.2 &75.4 & 85.1\\
%$\text{HierLSTM}_{MM}$  &   &   &    &     &  &  &  &  &  &  &  & X & & & &\\
% $\text{HierLSTM}_{MM}$  &   &   &    &     & 66.4 & 59.3 & 60.6 & 93.0 &  &  &  &  & & & &\\
$\text{HierbiLSTM}_{MM}$ & 2 (RS+C)& 42.0  & 33.8  & 35.1   & 88.5    & 68.9 & 62.8 & 63.2 &  93.5 & 84.1 & 80.8 & 81.3 & 94.5 & 82.3&77.9 &78.8 & 87.8 \\
Final & 3 (All)&  41.0 & 33.3  &  34.3  & \textbf{89.7}    & 71.4 &  62.7 & 63.2 & \textbf{93.8}  & 85.5 & 81.2 & 82.6 & \textbf{94.8} & 84.1 &78.3  & 80.2  & \textbf{88.7} \\
\end{tabular}
}
\caption{Results of the end-of-season classification models with different modalities (Remote Sensing, Crop Rotation, and Spatial Crop Distribution).
% combinations of Remote Sensing (RS), Crop Rotation (C) and Spatial Crop Distribution (D) . 
The metrics shown are Macro precision, recall and F1 score, as well as accuracy and micro-F1 score (m-F1).}
\label{tab:eof_classif}
\vspace*{-.4cm}
\end{table*}

\vspace*{-.2cm}
\subsection{Experimental protocol}

We trained all the networks via mini-batch stochastic gradient descent using Adam as optimizer \cite{Kingma2014} with a learning rate of $10^{-3}$ and a cross-entropy loss function. The number of neurons for the crop embedding layer, both the RNN internal layers, and the fully connected RS layer $f_{rs}$ as well as the number of stacked LSTM were chosen using hyperparameters search. The sizes of the layers $f_{c1}$ and $f_{c2}$ are the same than the one from the second RNN state $\textbf{h}_t$.

We trained our networks as for a sequence classification task, always with ten years of data. The labels from 2018 were used as training set, while the labels from 2019 as development set and the labels from 2020 as test set. All results presented hereafter refer to the analysis of 2020 crop types, which are based on models trained with the period 2009-2019, thus independent from the 2020 crop types observations. We zero-padded when no RS data was available (before 2016).

We proceed to a data-augmentation for the in-season classification model by cropping randomly the end of the timeseries for each batch starting from mid-March.   
All models were coded using the PyTorch library \cite{paszke2019pytorch}.% and all of our code will be available online after the reviewing period (\redcom{ok pour supprimer ca en attendant qu on fasse une 2eme publication ?} )

%\section{Results (Valentin Martin)}

\vspace*{-.2cm}
\subsection{Results}

%The data set includes 386 distinct cultivated code items. Knowing, the focus of this study on crop types, a set of land use and land cover categories was aggregated by an expert from the domain.
% a set of land use and land cover categories was aggregated and selected with respect to the aims and objectives of the min- istry.
%Hence, we propose several aggregation levels yielding to 28 and 12 classes as shown in Figure \ref{fig:distribcrop}.

In this Section we will show the results with two different settings: the classical setting where the network sees the whole year of RS signal, and a special early-season setting where the RS signal of the current season stops before the end of the season. % Our best model was obtained with a crop embedding size of 64,  
In order to deal with unbalanced classes, we used unweighted 
%means 
F1, Precision and Recall 
as well as the Accuracy. We used also the micro-F1, which is equivalent to Accuracy when having removed classes% for the 10-class configuration).\footnote{Equivalent to Accuracy with removed classes.} 
.

We also present results for 10 classes, which is the 12-class settings without grassland and other crops (see Figure \ref{fig:distribcrop}).

\vspace*{-.3cm}
\subsubsection{End-of-Season Classification} 

%\redcom{Finir quand le tableau est rempli}

The results of the end-of-season classification are available in Table \ref{tab:eof_classif}. We tested different configurations of networks, using different kind of features. The best results are obtained with our final model using information from the crop rotations, the S2 time series and the crop distribution of the surrounding fields.\footnote{from 2019} 

At first glance, we can see that the model using only the crop rotations can still reach an Accuracy of 73.3\% for the 386-class problem even if it does not use any information from the current year to make it's prediction.%, since it is purely made like a language model. 

Our RS models reach high results on the 386-class (up to 78.7\% with the $\text{HierbiLSTM}_{RS}$ model) due to the fact that, contrary to the main part of the works, they also use RS data from the past years on the same parcel, allowing to model a temporal context.  Interestingly, the Hierarchical setup with RS only allows for reaching higher results on the 386-class configuration, going from an accuracy of 72.5 to 78.7, when compared to the $\text{LSTM}_{YI}$.  

Finally, the local crop distribution vector allow for a slight improvement, which is more visible in the 10-class configuration. However, it unexpectedly decreases the macro-F1 while increasing the Accuracy for the 386-class configuration. This can be interpreted as the model making more mistakes on non-frequent crops only because it's globally better.
%One 
An explanation can be that the non frequent crops are not all situated in the same area, hence their distribution probability density is always approximated as 0. 

\vspace*{-.3cm}
\subsubsection{Toward In-Season Classification}

We saw earlier that the RS signal has shown pretty good end-of-season results, but it is known that the performances are strongly degraded when classifying during the season%, especially for some crops like the winter crops 
\cite{Russwurm2019b}. In this case, the crop rotations enhanced modality can help. 

\begin{figure*}%[!h]
    \vspace*{-.5cm} \hspace*{-.5cm}
    \centering 
    \includegraphics[width=1.08\textwidth]{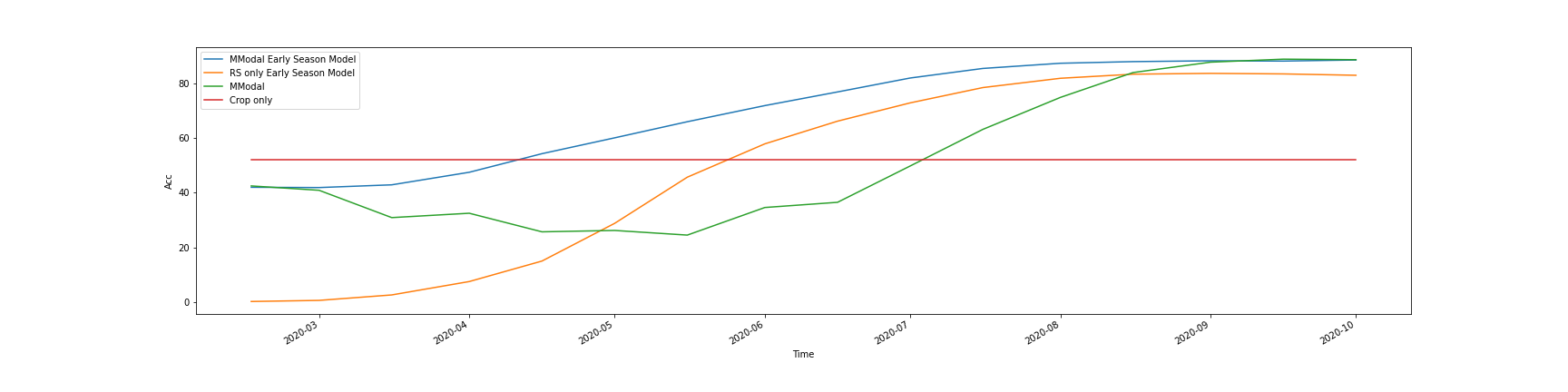} \vspace*{-.5cm}
    \caption{Comparison of Early classification using different modalities, with/out data augmentation (m-F1 with 10 classes).} \vspace*{-.5cm}
    \label{fig:early} 
\end{figure*}

For the in-season classification, we simply used our model trained over the whole year with data stopping at a point of the year. In Figure \ref{fig:early}, we compared the model using RS signal only with the multimodal model. It is important to notice that we used the same "final" model to adapt our domain to this noisy setup. The missing features, corresponding to unused months, were replaced by zeros. The results are thus preliminary and it is expected to obtained poor performances. A straightforward option could be to train new models for each of the evaluated months of the in-season classification. %je prefere mettre cette phrase en conclu. a toi de voir

The multimodal model always outperforms the RS model which is expected, especially at the beginning of the season when almost no information is available using the RS modality.

% Mettre un autre tableau avec les resultats en fonction du temps? 
%Figure \ref{fig:early}

%%%%%%%%% deleted to gain some space %%%%%%%%%s 
% \begin{table}%[H]
%     \centering
%     \resizebox{.42\textwidth}{!}{
%     \begin{tabular}{l|rrrr}
%     \textbf{Crop} &   \textbf{FAPAR} &     \textbf{LAI} &     \textbf{B04} &     \textbf{B8A} \\
%     \hline
%     Beetroot             &  99.994 &  99.999 &  99.990 &  99.998 \\
%     Consumption potatoes &  99.792 &  99.959 &  99.769 &  99.933 \\
%     %Grassland             &  99.970 &  99.996 &  99.977 &  99.994 \\
%     Maize                &  99.701 &  99.938 &  99.512 &  99.823 \\
%     Seed potatoes        &  99.888 &  99.886 &  99.973 &  99.931 \\
%     Sillage maize        &  99.863 &  99.893 &  99.939 &  99.876 \\
%     Spring wheat         &  99.981 &  99.997 &  99.979 &  99.988 \\
%     Starch potatoes    &  99.385 &  99.786 &  97.683 &  99.637 \\
%     Summer barley        &  99.439 &  99.698 &  99.767 &  99.624 \\
%     Winter barley        &  99.779 &  99.913 &  99.396 &  99.927 \\
%     Winter wheat         &  99.987 &  99.988 &  99.963 &  99.978 \\
%     \end{tabular}
%     }
%     \caption{Correlations between the actual and predicted RS time series.}
%     \label{tab:corr}
%     \vspace*{-.3cm}
% \end{table}
% Beetroot, Consumption potatoes, Maize, Seed potatoes, Sillage maize, Spring wheat, Starch potatoes, Summer barley, Winter barley, Winter wheat

\vspace*{-.2cm}
\section{Analysis}
\vspace*{-.1cm}

For the sake of clarity, all analyses presented hereafter in this section are limited to a set of crops of interest, corresponding to the 10-class setting. Details on the crops are provided in Figure \ref{fig:distribcrop}.
%\footnote{Beetroot, Consumption potatoes, Maize, Seed potatoes, Sillage maize, Spring wheat, Starch potatoes, Summer barley, Winter barley, Winter wheat.% Grassland and Other Crops not taken into account in the m-F1 calculation
%}

\vspace*{-.2cm}
\subsubsection*{High Precision examples}

\begin{table}%[]
\begin{tabular}{l|lll|lll}
 \textbf{Labels} & \multicolumn{3}{c}{\textbf{12-class}} & \multicolumn{3}{c}{\textbf{10-class}} \\ %\hline
  \textbf{Thresh}  & P & F1 & Acc & P & F1 & m-F1  \\ \hline \hline
$none$   & 85.3  & 82.6  &  94.8  &  84.1   & 80.2 & 88.7 \\
%   .8    &   &   &    &     &  &  &  & & \\
   .9   & 93.7   & 88.2    &  98.2   & 93.1   & 86.4    &  95.8
\end{tabular}
\caption{Results with the model using only the examples with a high probability for the predicted class.}
\label{tab:thresh}
\vspace*{-.5cm}
\end{table} %\vspace*{-.5cm}

We are presenting the results of our model on a fewer parcels where the precision is better than normal. In the perspective of crop monitoring, this analysis can be very valuable. Even if not 100\% of the parcels are aggregated, the output might support crop yield forecasting system, through the analysis of the crop specific RS time series with highest probability. %Note that the probability is an output of the model and is not based on 2020 validation set. 

We are taking the examples that are classified with a probability superior to 0.9 and compute some metrics over them. Those examples represent a big part of the dataset, they are more than 536k for the 12-class dataset and more than 148k for the 10-class dataset, representing respectively 90.0\% of all the parcels and 76.5\% of the parcels containing crop of interests. The results are shown in Table \ref{tab:thresh}. 

%%%%%%%%% deleted to gain some space %%%%%%%%%
%The RS time series of the main crops are analyzed at the country scale (Figure \ref{fig:visu}). The phenology is clearly identified, showing divergence between summer and winter crops, and correlated signals (B8A, LAI, FAPAR) opposed to B4. The predicted curves are perfectly matching the observed ones and correlations are very high (in Appendix, Table \ref{tab:corr}). This comfort the usefulness of the entire study to predict large scale crop specific RS signal.

%%%%%%%%% deleted to gain some space %%%%%%%%%
% \vspace*{-.3cm}
% \subsubsection*{Attention Visualization}
% \vspace*{-.1cm}

\vspace*{-.2cm}

\subsubsection*{In-Season Classification}%\vspace*{-.2cm}

We compare the vanilla model with the in-season classification model trained with our data-augmentation technique. The vanilla model has only seen during training examples of end-of-season classification, it is normal that they perform worst when used in in-season. This explains the fact that there is a decrease in performance compared to a model only taking into account the crops.% rotations and not the RS.

The data-augmentation used for the in-season models surprisingly does not work with RS only model, but allow the multimodal model to overpass the crop-only model in April. 

It is also interesting noting that the performances go below the unimodal crop model. This is certainly related that the models may give too much attention to the RS modality compared to the other ones, because the RS data modality has higher impact on the performance as the season progresses. An option to counter this effect would be to use a gate that would discard a noisy modality, as shown in \cite{Arevalo2017,chen2017multimodal}.

\vspace*{-.3cm}
\section{Conclusion and Future works}

We presented an innovative study to produce in-season crop mapping without relying on in-situ data of the current season. The approach relies on the analysis of several modalities, including the crop rotation of the  previous years, the Sentinel-2 time series of previous and current year as well as the previous year local crop distributions in the neighborhood parcels. 
A deep learning algorithm was used to model all those modalities at different level using a Hierarchical LSTM model. Firstly, we modeled the RS data with a Bidirectional-LSTM with Attention, using a sliding window on the satellite signals and integrating them using statistical functionals as it can be done for speech. Secondly, we fed the representation into another LSTM network modeling the crops as words and their rotation as sentence as it can be done with a language model. Finally, we added a context vector on the last layer in order to add information about the geographical place of the parcel. The designed methodology was tested over cropland of the Netherlands, benefiting from 12 years of crop rotation data nationwide.
%We obtain, for instance, a 97.9\% precision on winter wheat when using 91.4\% of the full observation. This offers new opportunities for crop areas counting and crop yield modeling at large scale, these two components being essential to forecast the crop production.
More generally, our method outperforms by a great margin the classical state-of-the-art using only a RNN or a Transformer to model the EO data at the level of a year. 

Nevertheless, there is still a lot of place for future work. More spectral bands added in the EO data could improve the performances. A better way to model the multimodality, at the level of EO data using multimodal aligned or non-aligned time-series fusion models\cite{Zadeh2018a,Yang2020mtgat}, and at a higher level between static representations \cite{Arevalo2017}. Finally our model impossible to adapt to an unknown place where the crop rotations are not available, a domain adaptation method using few-shot learning could be useful in this case \cite{Russwurm2019meta}. 
\bibliography{JRC_CVPPA.bib}

%%
%% If your work has an appendix, this is the place to put it.
\appendix

% \section{Online Resources}

\end{document}